\documentclass[oneside]{article}

\usepackage{PRIMEarxiv}

\usepackage{times}
\usepackage{url}
\usepackage[hidelinks]{hyperref}
\usepackage[utf8]{inputenc}
\usepackage[small]{caption}
\usepackage{graphicx}
\usepackage{amsmath}\usepackage{amsfonts}
\usepackage{amsthm}
\usepackage{booktabs}
\usepackage{algorithm}
\usepackage{algorithmic}

\usepackage{makecell}
\usepackage{multirow,tabularx}
\usepackage{array}
\newcolumntype{L}[1]{>{\raggedright\let\newline\\\arraybackslash\hspace{0pt}}m{#1}}
\newcolumntype{C}[1]{>{\centering\let\newline\\\arraybackslash\hspace{0pt}}m{#1}}
\newcolumntype{R}[1]{>{\raggedleft\let\newline\\\arraybackslash\hspace{0pt}}m{#1}}
\title{CloudNine: Analyzing Meteorological Observation Impact on Weather Prediction Using Explainable Graph Neural Networks}
\author{
Hyeon-Ju Jeon, Jeon-Ho Kang, In-Hyuk Kwon\\
Data Assimilation Group\\
Korea Institute of Atmospheric Prediction Systems\\
Seoul, Republic of Korea\\
\texttt{hjjeon,jhkang,ihkwon@kiaps.org}\\
\And
O-Joun Lee${}^{\dagger}$\\
Dept. of Artificial Intelligence\\
The Catholic University of Korea\\
Bucheon, Republic of Korea\\
\texttt{ojlee@catholic.ac.kr}\\
}

\begin{document}
\maketitle

\begin{abstract}
The impact of meteorological observations on weather forecasting varies with sensor type, location, time, and other environmental factors. 
Thus, quantitative analysis of observation impacts is crucial for effective and efficient development of weather forecasting systems. 
However, the existing impact analysis methods are difficult to be widely applied due to their high dependencies on specific forecasting systems. 
Also, they cannot provide observation impacts at multiple spatio-temporal scales, only global impacts of observation types. 
To address these issues, we present a novel system called ``CloudNine,'' which allows analysis of individual observations' impacts on specific predictions based on explainable graph neural networks (XGNNs).
Combining an XGNN-based atmospheric state estimation model with a numerical weather prediction model, we provide a web application to search for observations in the 3D space of the Earth system and to visualize the impact of individual observations on predictions in specific spatial regions and time periods. 

\let\thefootnote\relax
\footnotetext{${}^{\dagger}$ Correspondence: \texttt{ojlee@catholic.ac.kr}; Tel.: +82-2-2164-5516} %%%%%%%%%%
\end{abstract}


%관측이 예측에 미치는 영향을 정량적으로 분석하여 정확한 기상 예보를 위한 관측 전략을 개발하고, 자원을 효과적으로 할당하여 예측 정확도를 향상시키는 데 기여합니다. 이는 다양한 분야에서 안전성, 효율성, 경제성을 향상시키는 데 도움이 됩니다.
\section{Introduction}
% 관측 종별로 볼 수 있지만, 개별 관측이 지역적으로 미치는 영향을 검색할 수 있는 시스템은 없다.

This system, ``CloudNine,'' aims to assess meteorological observations' impacts on weather forecasting with multiple spatio-temporal resolutions and visualizations. 
%Weather forecasting is one of the fields where the adoption of deep learning models has been slower than in other fields. 
Existing weather forecasting systems rely on numerical weather prediction (NWP) systems based on 3D physical models and dynamical equations \cite{Stulec2019,Kotsuki2019}. 
%Since NWP predicts future atmospheric states using simulations based on current states, existing deep learning models still struggle to emulate correlations between global meteorological variables with reasonable computational costs. 
%Therefore, we focus on applying deep learning models to other modules surrounding NWP. 
NWP predicts future atmospheric states using simulations based on current states.
In obtaining current atmospheric states, we cannot observe global atmospheric states with regular spatial grids and time intervals, and data assimilation (DA) systems play an important role in approximating atmospheric states by merging observations with prediction results from dynamical models \cite{Kwon2018}. 
Observations are obtained at irregular locations and times from various sources such as aircraft, radar, and satellites. 
Therefore, selecting appropriate observations from this variety improves the accuracy of DA and NWP \cite{Kwon2018}. 
Although there have been various methods for assessing observation impacts \cite{Kotsuki2019,Kalnay2012,Buehner2018}, they provide only global impacts of observation types, not of each observation for specific regions and periods, and cannot be widely used due to their high dependencies on specific NWP and DA systems. 

%\begin{figure}[t]
%	\centering
%	\includegraphics[width=0.75\columnwidth]{fig_example.png}
%	\caption{An example of observation impact visualization.}
%	\label{fig_overview}
%\end{figure}

To improve these problems, CloudNine employs explainable graph neural networks (XGNNs). 
Recently, GNNs have been widely used to predict meteorological variables, such as solar radiation and sea surface temperature, by capturing correlations between variables in adjacent regions \cite{Jeon2022,Ma2023,Yang2018,Hoang2023,Lee2021,Lam2023}. 
Our previous study \cite{Jeon2024} exhibited that GNNs are also effective in estimating current atmospheric states by fusing observations with spatial grids of NWP. 
Also, the previous study applied three explainability methods (e.g., SA \cite{Pope2019}, Grad-CAM \cite{Pope2019}, and LRP \cite{baldassarre2019explainability}) to observation impact analysis. 
CloudNine employs only LRP among the three methods, which showed the best performance. 
Then, CloudNine conducts impact analysis at multiple spatio-temporal resolutions by aggregating the impacts according to observation types, spatial regions, and time periods. 
Finally, CloudNine provides visualization tools to assist users in understanding the multi-resolution impact analysis to let them select meteorological variables effectively and efficiently in designing forecasting systems. 
A demonstration video of CloudNine is available at web\footnote{\url{https://youtu.be/vqhk6fCCGR0}}.
The contributions of this paper include:
\begin{itemize}
    \item We developed CloudNine based on XGNNs for atmospheric state estimation and observation impact analysis without dependencies on specific NWP and DA systems.
    \item CloudNine enables impact analysis of individual observations for predictions at a certain location and time. 
    \item CloudNine includes visualization tools for observation impact analysis at multiple spatio-temporal resolutions. 
\end{itemize}

\begin{figure}[t]
	\centering
	\includegraphics[width=0.9\columnwidth]{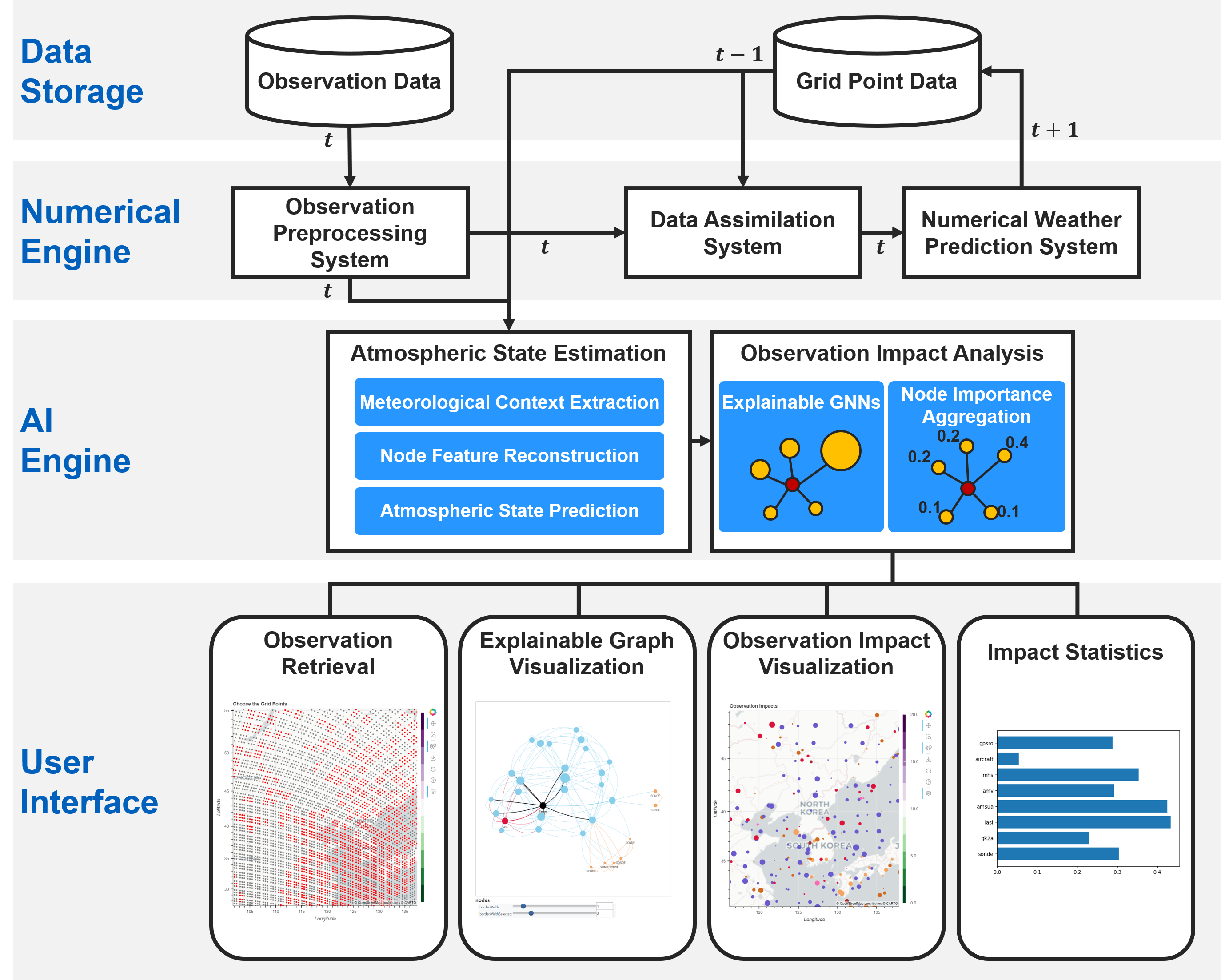}
	\caption{An overview of the proposed system, CloudNine.}
	\label{fig_overview}
\end{figure}

\section{System Design}

We present an overview of the ``CloudNine'' system in Figure \ref{fig_overview}. 
%, where pre-processed real-time observation data and the latest grid point data are fed as inputs, and the AI engine based on XGNNs visualizes observation impacts as output while predicting the current atmospheric state.
The system consists of four components. 
First, the numerical engine pre-processes real-world observations, assimilates the pre-processed observations and grid point data (i.e., atmospheric states predicted by NWP based on previous observations), and predicts atmospheric states in the next time steps.
%Then, the AI engine extracts the meteorological context and relevance between the grid point data and observations.
Then, the AI engine extracts meteorological contexts and analyzes the observation impacts by using XGNN methods.
Finally, the user interface facilitates the visualization of observation points and observation impacts, providing the model's insights comprehensibly.
%We will introduce each component in detail below.
%The workflow of the ``CloudNine'' is shown in Figure \ref{fig_overview}. 
%The proposed system consists of four components: data storage, numerical engine, AI engine, and user interface. 

\subsection{Meteorological Dataset}
%The dataset for AI engine comprises of data from NWP grid points at time \( t-1 \) and observations at time \( t \).
% Numerical Engine의 결과물을 AI Engine으로 분석하기 위해서는 해당 자료를 바탕으로 dataset을 만들어야한다.
% 

The inputs to the AI engine consist of observations at time $t$ and data from NWP grid points, which are atmospheric states at $t$ predicted by NWP based on atmospheric states at $t-1$. 
%Simply speaking, NWP is a recursive model, and our system composes inputs of NWP at $t$ by fusing its outputs at $t-1$ and new observations at $t$. 
%
CloudNine includes a function to build an empirical atmospheric dataset.
Recently, global temperature change has affected atmospheric circulation patterns, resulting in various extreme weather events that had barely occurred in the past. 
To handle this problem, it is essential to build a dataset by collecting the latest observational data and NWP-modelled atmospheric data. 
Therefore, we collected the observations and NWP grid point data from the Korea Meteorological Administration (KMA), which operates a practical weather forecasting system.

As in our previous study \cite{Jeon2024}, we represent meteorological correlations between spatially adjacent locations as graphs. 
In the Korean Integrated Model (KIM), observations influence NWP grid points within a 50km radius, which is the same in CloudNine. 
A meteorological graph ($\mathcal{N}=(V, E,\lambda_v)$) has observation points and NWP grid points as nodes ($V \ni v_i$) and spatial adjacencies as edges ($E \ni e_{ij}$). 
%The meteorological graph based on their spatial adjacencies can effectively fuse them by analyzing their correlations. 
%
%The definition of meteorological graphs was discussed in detail in our previous study\cite{Jeon2024}.
%In this study, for system development, global weather data is divided into four regions to prevent data overload when visualizing observation and NWP grid points. 
In this study, we divide the global meteorological data into four regions to avoid data overload in the user interface.
The data set is divided into Asia, Europe, North America, and Australia according to the WMO latitude and longitude ranges and includes about 500 hPa pressure levels.
The NWP grid points have node features, including four meteorological variables, such as u-component of wind (U), v-component of wind (V), temperature (T), and relative humidity (Q).
Among various observation sources in KMA, 11 sources (e.g., AIRCRAFT (U, V, T), GPSRO (banding angle (BA)), SONDE (U, V, T, Q), AMV (brightness temperature (TB)), AMSU-A (TB), AMSR2 (TB), ATMS (TB), CrIS (TB), GK2A (TB), IASI (TB), and MHS (TB)) are acquired for our dataset.

\subsection{Weather Prediction}

The atmospheric state prediction model based on GNNs is designed to analyze spatial correlations between meteorological phenomena and to extract distinct characteristics of each observation. 
This model is combined with a gradient flow-based explainable method, LRP \cite{baldassarre2019explainability}.
%The GNN-based weather prediction model for understanding the spatial dependency of meteorological phenomena and extracting the different characteristics of each observation species is combined with a gradient flow-based explainable method to form an AI engine. 
%For analyzing the observation impact, we use GNN-based node-level regressor combined with gradient flow-based explainable method.% in which we first collect data
The weather prediction model uses data from NWP grid points at time $t-1$ and observations at time $t$ to predict atmospheric states of the grid points at time $t$.
We perform a node feature reconstruction task to make the model understand different types of observations and obtain node feature vectors that represent distinctive characteristics of observations and grid points. 
%to update node features that represent the distinctive characteristics of nodes.
Given the diverse sensing methods and units for each type of observation, this task is essential for handling a wide range of observation modalities.
Then, a node-level regression task is performed to predict the current atmospheric states at each grid point.
% 관측 종별로 서로다른 특성을 학습하기 위해서, node freature reconstruction을 수행하여 node의 구별되는 특성을 표현할 수 있는 벡터로 node feature를 업데이트 한다. (관측마다 센싱하는 방식과 단위가 상이하므로, 다양한 modality의 관측을 다루기 위해서는 이 과정이 필수적이다.)
% 그 다음에 node-level regression 과정을 거쳐서 각 그리드 포인트의 값을 예측한다.
%
%To assess the impact of new observations, their neighboring grid points are extracted, and the meteorological contexts for each grid point are composed.
%Features of each node in the meteorological contexts are reconstructed, capturing meteorological correlations between neighboring grid points and the new observations.
%Finally, node-level regression is performed using the updated node features, and the current atmospheric state of the target grid point is then output.
% 새로운 관측의 영향을 평가하고 싶다면, 해당 관측이 영향을 미치는 인접 그리드 포인트를 추출하고, 각 그리드 포인트 마다 meteorological context를 생성한다. 
% 그 다음에 각 meteorological context에서 각 노드의 feature가 reconstruction 되며, 이 과정을 통해 인접 그리드 포인트 간의 연관관계와 관측들의 구별되는 특성들이 학습된다. 
% 마지막으로, 학습된 노드 feature를 바탕으로 node-level regression을 수행하면, 타겟 그리드 포인트의 현재 대기 상태가 output이 된다. 

\subsection{Observation Impact Analysis}

We use a widely-used explainable method to estimate observation impact.
%In our previous study \cite{Jeon2024}, we compared three gradient-based graph explainable methods (e.g., SA \cite{Pope2019}, Grad-CAM \cite{Pope2019}, and LRP \cite{baldassarre2019explainability}) and validated their ability to estimate the impact of observations in the Asian region.
In our previous study \cite{Jeon2024}, we compared three gradient-based graph explainable methods (e.g., SA \cite{Pope2019}, Grad-CAM \cite{Pope2019}, and LRP \cite{baldassarre2019explainability}) and validated their ability to estimate the impact of observations.
In this study, we evaluate and analyze the impact of observations using the layer-wise relevance propagation (LRP) method, which showed the most consistent performance.
The explainable method involves back-propagating prediction results from the output of GNN layers to the input features. 
This method assigns a higher score to the connecting neurons that contribute more to the activation of the target neuron.

%Given a target grid point, our system constructs a subgraph called meteorological contexts that includes neighboring observations and grid points and predicts atmospheric states at the target node.
%Then, the system applies an explainable method to reveal the impact of each node on the predicted atmospheric state.
To assess the node's importance, considering that one observation may belong to multiple meteorological contexts, we aggregate the importance from multiple meteorological contexts associated with a single observation to produce an average impact.
The spatio-temporal range of aggregation can be easily adjusted according to purposes of impact analysis. 
As a result, the user interface can visualize and analyze the impact of each observation on neighboring areas, providing valuable insights to meteorology experts in designing weather forecasting systems.

\section{Results and Demonstration}

\subsection{Visualization with Interactive Web Application}

\begin{figure}[t]
	\centering
	\includegraphics[width=0.73\columnwidth]{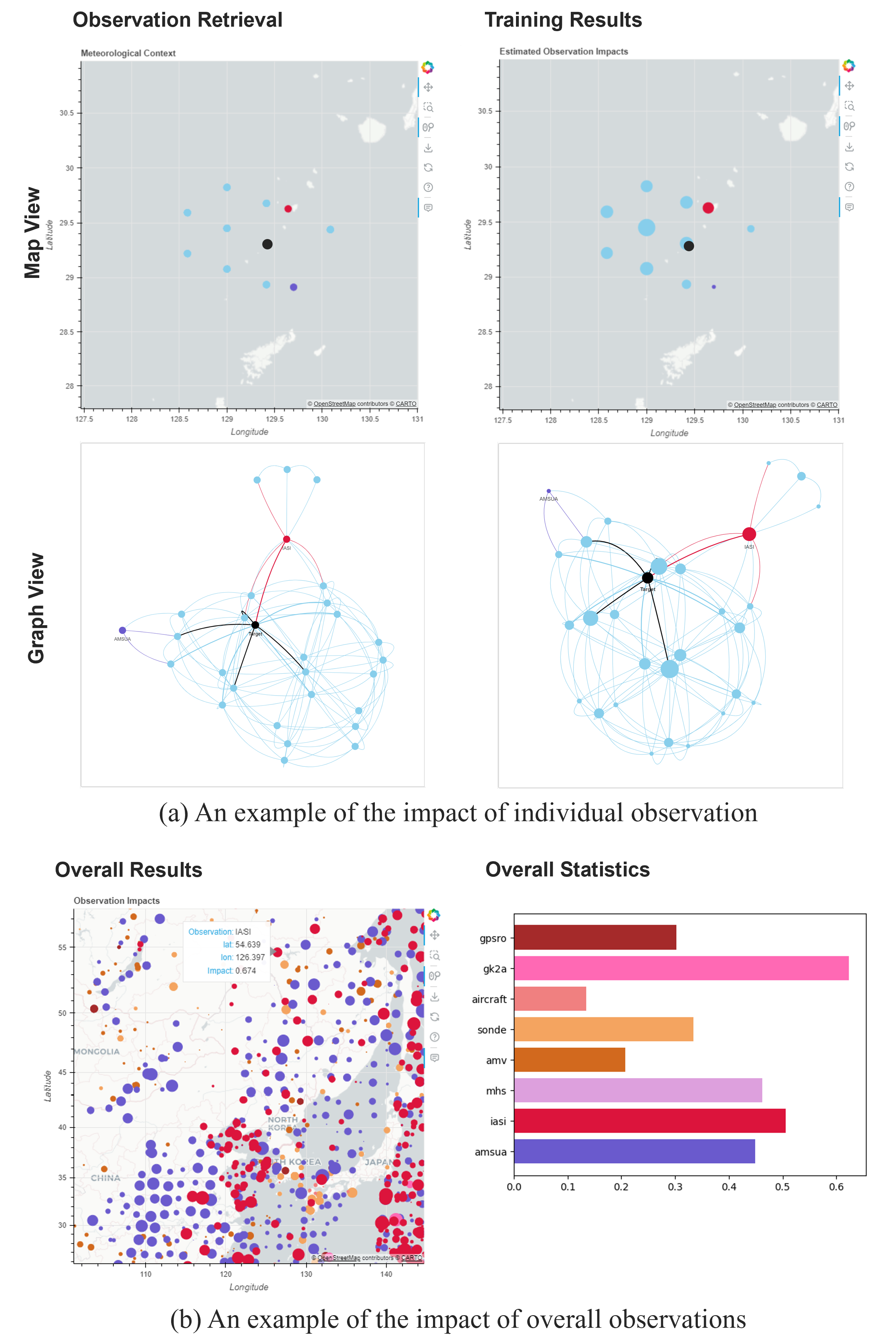}
	\caption{Examples of visualizing observation impacts by using the interactive web application. Each point indicates nodes in the meteorological graph. The node color represents the types of nodes, and the node size refers to the node's importance. }
	\label{fig_demo}
\end{figure}

Figure \ref{fig_demo} presents examples of visualization results of CloudNine.  
To handle the aforementioned modules and to explain the node importance for newly added observations, we build the system with a common front-end and back-end architecture.
%This separation reduce the possibility of encountering issues related to security and data access.
By maintaining this separation, the possibility for encountering issues related to security and data access is significantly reduced. 
Also, the architectural structure improves the adaptability of the system, as the back-end can be deployed on the outsourced system to facilitate scalability and enhance system performance.

The back-end is built in Python and utilizes a Flask framework to communicate with the front-end.% via an IP network. 
The project is saved as a folder in the platform's file system containing the dataset and the trained model.
The system accepts file formats of .nc for the dataset and .pt files for the model. 
All processing is performed in the back-end to minimize the performance demands on the front-end.
The front-end utilizes the VisJS and BohkeJS libraries as our main components for visualizing observation data with interactive plots.
%The front-end employs the VisJS and BohkeJS libraries for our primary components to visualize observation data with interactive plots. 
%All together form a responsive and fluently scaling web application.

% figure !!!! -> prediction model 구조 그림도 있음 좋겠다
%The workflow of the CloudNine consists of four parts:
%(1) Starting with the setup process, the back-end retrieves project information from the file system and forwards it to the front-end.
%(2) The dataset preparation process begins, where the observation pre-processing system and the data assimilation system are run using real-time observation data and grid point data.
%(3) After the user selects the target region and date to be analysed, the preparation process starts by loading the trained model and propagating the weight of the model's neuron.
%(4) Finally, the user can interact with the system by estimating the impact of newly added observations on adjacent grid points through the loaded model layers. 

\subsection{Effectiveness of Observation Impact Analysis}
%정확도 표 요약본을 넣고, aaai workshop paper를 reference로 사용 가능함. 
%In this section, we evaluate the performance of the XGNN method based on prediction accuracy and consistency.
%Also, we discuss whether our CloudNine system is effective and useful to users based on the user interface.

\begin{table}[t]
	\centering
	\scriptsize
	\begin{tabular}{c|l|C{0.7cm}C{0.7cm}C{0.7cm}|C{0.7cm}C{0.7cm}}
		\toprule
        \multirow{2}{*}{\textbf{Region}}&\multirow{2}{*}{\textbf{Variables}}&\multicolumn{3}{c}{\textbf{Prediction}} &\multicolumn{2}{|c}{\textbf{Explainability}} \\
		& & {RMSE}& {MAE}& \textbf{$ACC$}& \textbf{$Fi.+$}& \textbf{$Fi.-$}\\
		\midrule
		\multirow{4}{*}{Asia}		
		& U(m/s)	& \textbf{0.09}&\textbf{0.07}&\textbf{0.82}&\multirow{4}{*}{\textbf{0.23}}&\multirow{4}{*}{\textbf{0.07}}\\
		& V(m/s)	& 0.10&0.08&0.82&&\\
		& T(K)		& 0.16&0.13&0.82&&\\
		& Q(kg/kg)	& 0.08&0.08&0.85&&\\
		\midrule
		\multirow{4}{*}{Europe}		
		& U(m/s)	& 0.10&0.08&0.78&\multirow{4}{*}{0.20}&\multirow{4}{*}{0.09}\\
		& V(m/s)	& 0.13&0.10&0.74&&\\
		& T(K)		& 0.19&0.17&0.80&&\\
		& Q(kg/kg)	& 0.12&0.12&0.78&&\\
		\midrule
		\multirow{4}{*}{\makecell{North \\America}}		
		& U(m/s)	& 0.11&0.08&0.78&\multirow{4}{*}{0.21}&\multirow{4}{*}{0.08}\\
		& V(m/s)	& 0.09&0.06&0.82&&\\
		& T(K)		& 0.16&0.14&0.83&&\\
		& Q(kg/kg)	& \textbf{0.04}&\textbf{0.04}&\textbf{0.92}&&\\
		\midrule
		\multirow{4}{*}{Australia}	
		& U(m/s)	& 0.13&0.10&0.75&\multirow{4}{*}{0.22}&\multirow{4}{*}{0.07}\\
		& V(m/s)	& \textbf{0.08}&\textbf{0.06}&\textbf{0.87}&&\\
		& T(K)		& \textbf{0.15}&\textbf{0.11}&\textbf{0.84}&&\\
		& Q(kg/kg)	& 0.07&0.07&0.88&&\\
		\bottomrule
	\end{tabular}
	%}
\caption{Effectivenss of the observation impact analysis.}
\label{table1}
\end{table}

Table \ref{table1} shows the prediction performance and explainability of CloudNine for each atmospheric variable by region.
All variables have consistent accuracy across all regions in each evaluation metric \cite{Zhao2020}.
The prediction accuracy of horizontal wind (U) and vertical wind (V) were the highest in Asia and Australia, respectively. 
Humidity (Q) and temperature (T) had the highest accuracy in North America and Australia, respectively. 
The European region exhibits the lowest prediction accuracy across all variables.
This comes from that Europe with its relatively small geographical scope has a smaller number of observation samples than other regions.
Also, for evaluating the explainable methods, we occlude $20\%$ of the input features of the nodes with the top ($Fi.+$) and bottom ($Fi.-$) of node importance and then compare their fidelity scores.
Fidelity was the highest in Asia.
We assume that the AI engine implicitly reflects the characteristics of KIM, such as a large number of well-preprocessed observations around the Korean peninsula and the high forecasting accuracy over the area.
%is most accurate in North America, and  is most accurate in Australia.
%The European region has the lowest prediction accuracy across all variables. 
% 설명 가능한 방법의 디스커션 (4-5문장)
% 일관적인 성능을 보이고, 아시아지역에서 가장 정확도가 높은데, 범위가 가장 넓고 다양한 육지 범위가 넓기 때문
%Let $Fi.+$ and $Fi.-$ be the fidelity scores after occluding the input features of the nodes with the top and bottom $20\%$ of node importance, respectively. 

Figure \ref{fig_demo} shows the interactive user interface of the CloudNine system. 
The visualization consists of a map view showing spatial locations of observations and a graph view presenting relationships between the observations and NWP grid points.
In Figure \ref{fig_demo} (a), the impact of each observation on a particular NWP point can be analyzed.
The user can navigate to the NWP point through observation search tool and confirm its location and meteorological context at that point.
Then, the meteorological context is used as input to the trained model, and the importance of each node is visualized based on the model's prediction results.
In Figure \ref{fig_demo} (b), we visualize the impact of individual observations. 
The impact of observation is influenced by various factors, including geographical characteristics, meteorological phenomena, and the impact of neighboring observations. 
Also, even observation types that have a high global impact may have a low impact at specific spatial regions or time periods. 
Therefore, visualization tools that can analyze the impact of individual observations play an important role.

% 

%\subsection{User Evaluation}

\section{Conclusion}
%Summary

This study proposes CloudNine, which can assess and visualize meteorological observations' impacts on weather forecasting. 
Although CloudNine is designed to be independent of specific weather forecasting systems, it currently imports only KIM prediction results. 
Comparing observation impacts on different NWP systems can provide more analytical insights. 
Therefore, our future study aims to integrate the prediction results of different weather forecasting systems. 

%The system can only accept the dataset from the operational weather prediction system of the Korean Meteorological Administration (KMA).
%Although multiple operational centres and researchers can produce better weather prediction results, our system shows the impact of observations on KMA outputs.
%Although there are several operational center and different weather prediction results, the system shows the observation influence on the KMA results.
%Future work:
%In our further study, we try to develop the system to accept the different prediction results as input and process and update the AI engine.
%This will be helpful for the expert group as they will be able to validate that their assumption contributes to the usefulness of the observations.

%\appendix

%\section*{Ethical Statement}

%There are no ethical issues.

\section*{Acknowledgments}

This work was supported 
in part by the R\&D project “Development of a Next-Generation Data Assimilation System by the Korea Institute of Atmospheric Prediction System (KIAPS)”, funded by the Korea Meteorological Administration (KMA2020-02211) (H.-J.J.)
and in part by the National Research Foundation of Korea (NRF) Grant funded by the Korea government (MSIT) (No. 2022R1F1A1065516 and No. 2022K1A3A1A79089461) (O.-J.L.).

%% The file named.bst is a bibliography style file for BibTeX 0.99c
\bibliographystyle{unsrt} 
\bibliography{arxiv_24_cloudnine}

\end{document}